\documentclass[letterpaper, 12 pt, journal, twoside]{ieeetran}
\IEEEoverridecommandlockouts
\usepackage{cite}
\usepackage{amsmath,amssymb,amsfonts}
\usepackage{algorithmic}
\usepackage{algorithm}
\usepackage{tabularx}
\usepackage{hyperref}
\usepackage{graphicx}
\usepackage{textcomp}
\usepackage{xcolor}
\usepackage{verbatim}
\usepackage{adjustbox}
\usepackage{multirow}
\usepackage{float}
\usepackage{booktabs}
\usepackage{fancyhdr}

\def\BibTeX{{\rm B\kern-.05em{\sc i\kern-.025em b}\kern-.08em
    T\kern-.1667em\lower.7ex\hbox{E}\kern-.125emX}}
\begin{document}

\markboth{UNDER REVIEW BY A ROBOTICS JOURNAL}
{UNDER REVIEW BY A ROBOTICS JOURNAL}

\author{Tianze Chen$^{1}$, Ricardo Frumento$^{1}$, Giulia Pagnanelli$^{2}$, Gianmarco Cei$^{2}$, Villa Keth$^{1}$, \\ 
Shahaddin Gafarov$^{1}$, Jian Gong$^{1}$, Zihe Ye$^{3}$, Marco Baracca$^{2}$, Salvatore D'Avella$^{4}$, \\
Matteo Bianchi$^{2}$ and Yu Sun$^{1}$
\thanks{$^{1}$ The authors are from the Robot Perception and Action Lab (RPAL) of Computer Science and Engineering Department, University of South Florida, Tampa, FL 33620, USA. Email: \texttt{\{tianzechen, ricardod, villak, gafarov, gongj, yusun\}@usf.edu}.}
\thanks{$^{2}$ The authors are with the Research Center "E. Piaggio", Department of Information Engineering, University of Pisa, Pisa, Italy. Email: \texttt{\{giulia.pagnanelli, gianmarco.cei, marco.baracca\}@phd.unipi.it, matteo.bianchi@unipi.it}.}
\thanks{$^{3}$ The author is with is with Rutgers University, New Brunswick, NJ 08901, USA. Email: \texttt{zihe.ye@rutgers.edu}. Related work was finished when Zihe Ye was a Master's student in the RPAL lab at USF. }
\thanks{$^{4}$ The author is with the Department of Excellence in Robotics \& AI, Mechanical Intelligence Institute, Scuola Superiore Sant’Anna, Pisa, Italy. Email: \texttt{salvatore.davella@santannapisa.it}.}
\thanks{Work partially supported by USF CREATE AWARD and the Italian Ministry of Education and Research (MUR) in the framework of the FoReLab project (Departments of Excellence)}.
}

\title{Benchmarking Multi-Object Grasping}

\maketitle

\begin{abstract}
In this work, we describe a multi-object grasping benchmark to evaluate the grasping and manipulation capabilities of robotic systems in both pile and surface scenarios. The benchmark introduces three robot multi-object grasping benchmarking protocols designed to challenge different aspects of robotic manipulation. These protocols are: 1) the Only-Pick-Once protocol, which assesses the robot's ability to efficiently pick multiple objects in a single attempt; 2) the Accurate pick-trnsferring protocol, which evaluates the robot's capacity to selectively grasp and transport a specific number of objects from a cluttered environment; and 3) the Pick-transferring-all protocol, which challenges the robot to clear an entire scene by sequentially grasping and transferring all available objects. These protocols are intended to be adopted by the broader robotics research community, providing a standardized method to assess and compare robotic systems' performance in multi-object grasping tasks. We establish baselines for these protocols using standard planning and perception algorithms on a Barrett hand, Robotiq parallel jar gripper, and the Pisa/IIT Softhand-2, which is a soft underactuated robotic hand. We discuss the results in relation to human performance in similar tasks we well.

\end{abstract}

\begin{IEEEkeywords}
Grasping, Benchmarking, Dexterous Manipulation, Multifingered Hands
\end{IEEEkeywords}

\section{Introduction} 

Robotic grasping is a key area of research in robotics and automation. Traditional robotic grasping methods typically handle one object at a time. If a task requires handling multiple items, the robot must repeat the process for each one. In contrast, humans can naturally grasp several items simultaneously. For example, a person might grab a handful of apples from a basket or multiple screws from a tray, depending on the need and what their hand can hold. In logistics and manufacturing, workers often perform batch picking, picking up several small items or containers and putting them into a bin to speed up the processes. However, enabling robots to grasp and hold multiple objects poses significant challenges. This requires more sophisticated techniques in perception and planning.

Grasping tasks could have multiple scenarios. Among them, surface picking and pile picking are the most common. Surface picking involves grasping objects from a flat surface, while pile picking deals with stacked objects. The flat surface could be a tabletop, a conveyor belt, or the bottom of a bin. The object pile could be in a bin, a bag, or a jar. The objects could be identical products of the same size, such as a box of the same candies or cosmetic jars with the same Stock Keeping Unit (SKU). They could also be the same organic object with the same shape but slightly different sizes, such as a bowl of eggs or a bag of apples. They could be a collection of different objects, such as a tabletop of random objects for tabletop arrangement tasks, a shopping basket of items, and a tote of items with different SKUs before re-binning in a warehouse. 

Several benchmarks have been proposed to measure the performance of robotic systems for picking one item at a time in many of those scenarios. In \cite{8936856}, a benchmark was designed for measuring the performance of picking and transferring a one-inch cube at a time from a pile of cubes in a box. In \cite{liu2021ocrtoccloudbasedcompetitionbenchmark}, a benchmark was developed to assess object rearrangement efficiency of picking and placing one item at a time. In \cite{8954746}, a benchmark was defined to measure the efficiency of picking and transferring a list of items from a tote, one item at a time. Several competitions have also been organized based on those benchmarks to measure and showcase the research development of single-object grasping in those scenarios \cite{7583659,sun2021researchchallengesprogressrobotic, liu2021ocrtoccloudbasedcompetitionbenchmark, 10271337}. The protocols and rules in those benchmarkers particularly forbid picking up multiple items at a time. 

In concept, many single-object-oriented grasping benchmarks could be expanded to measure the performance of grasping multiple objects as they may share the same goal: accurately picking a number of objects efficiently. Allowing robots to pick up multiple items at once opens up many opportunities for efficiency improvement. They naturally introduce many factors that are not counted for in those single-object-grasping benchmarks. For example, strategizing the grasping sequence becomes critically important, and grasping accuracy in one round may not be as critical. There is a clear need for a set of new benchmarks to unpack the influence of the factors and properly measure the progress of robotics research for multi-object grasping. 

\section{Related Work}
In addition to the benchmarking works on single-object grasping mentioned in the Introduction, we introduce here the related advancements in Multi-Object Grasping (MOG). Harada and Kaneko\cite{harada1998enveloping, harada2002active} explored multi-object grasping using robotic hands, focusing on enveloping grasp and active force closure (AFC). They analyzed kinematic conditions for stable grasping, ensuring rolling contacts at interfaces, and derived mathematical models to validate AFC feasibility. Later, \cite{yoshikawa2001optimization} optimized power grasps by maximizing stability while minimizing gripping force, validated via simulations. \cite{yamada2005grasp} derived stability conditions for grasping two objects in 2D based on geometry and contact points, while \cite{yamada2015static} extended this to 3D, modeling robotic fingers as linear springs and incorporating contact surface geometry. These studies collectively provide theoretical and simulation-based foundations for stable multi-object grasping.

Recent methods extend these ideas to more complex scenarios. \cite{https://doi.org/10.48550/arxiv.2206.00229} proposed a multi-object push-grasping approach for rigid convex polygonal objects on a planar surface. They established the necessary conditions for frictionless push-grasps and developed a grasp planner to filter inadmissible grasps. Expanding on this, \cite{10260295} introduced Push-MOG, which used "fork pushing" actions with a parallel-jaw gripper to cluster objects, facilitating multi-object grasping. Push-MOG employs hierarchical clustering to identify clusters and executes stable pushes to optimize grasp configurations. Other methods incorporate learning-based approaches. MOG-Net \cite{10341895} plans robust grasps for multiple convex polygonal objects by integrating frictional interactions. \cite{Yao_2023} proposed a method for multi-fingered robotic hands to grasp multiple objects by leveraging kinematic redundancy. Their algorithm uses pairwise contacts and a reachability map to optimize stable grasps, validated on a robotic hand. The work in \cite{9636777} tackled the problem of estimating the number of objects within a robotic grasp when the hand is submerged in a pile, highlighting the importance of precise perception. Similarly, \cite{9812388} analyzed human and robotic motion data to define 12 distinct grasping types, providing a structured foundation for improving grasp strategies based on real-world insights.

For scenarios involving transferring objects, the transferring algorithm based on Markov decision processes (MDP) \cite{9981799} ensures consistent and efficient execution. The Experience Forest algorithm \cite{10502193} enhanced robotic precision by organizing finger movements into structured patterns, enabling more effective interactions with objects. The Only-Pick-Once (OPO) system \cite{ye2023pick} demonstrated significant improvements in efficiency through optimized grasp-and-transfer processes. \cite{yonemaru2025learninggroupgraspmultiple} introduces an imitation learning-based approach using a diffusion policy network to enable robots to dynamically generate action sequences for pushing, grouping, and grasping multiple objects simultaneously. Advanced systems and designs have further elevated MOG performance. A multimodal gripper system integrating a suction cup was proposed by \cite{10049497}, allowing for enhanced flexibility in handling different types of objects.  Human-inspired designs like MOGrip \cite{doi:10.1126/scirobotics.ado3939} emulate natural grasping abilities, enabling sequential object handling, secure storage, and smooth transfer in a single operation, setting new standards for robotic dexterity and adaptability.

\section{Benchmarking Protocols}
Standardized benchmarking is vital for evaluating and comparing robotic systems in multi-object grasping, providing a framework to identify strengths, address weaknesses, and set clear performance goals. Consistent metrics and protocols enhance reproducibility and facilitate meaningful comparisons, driving the development of more robust and versatile systems for real-world applications. In this paper, we propose three benchmarking protocols to evaluate robotic systems' MOG capabilities in pile and surface scenarios with identical objects. The first protocol is the only-pick-once (OPO) protocol. It is designed to measure the performance of grasping multiple objects in a single attempt. The second protocol is the accurate pick-transferring (APT) protocol. It is developed on top of the OPO protocol and aims to measure the overall efficiency of picking and transferring a targeted number of objects. The last one is the pick-transferring-all (PTA) protocol. It is developed to measure the efficiency of picking and transferring all items from one container to another. In each of the protocols, we use both surface and pile scenarios with a robotic manipulator and a bin. 

\paragraph{Robot Setup} 
The experimental setup consists of a robotic manipulator positioned in front of a support surface, with an RGB-D camera mounted either above or on the robotic arm to enable real-time object detection and pose estimation. The camera captures depth and color data, which vision algorithms process to identify object locations and orientations within the source bin, facilitating grasp planning and execution. In some experiments, a simulated counterpart replicates the real-world setup, maintaining identical relative positions and dimensions to ensure consistency in benchmarking.

\paragraph{Object Setup and Scenarios}
Both the source and target bins are placed within the robot's reachable workspace, with their exact positioning left to the user’s discretion. The source bin contains objects arranged in either piles of identical items or a single-layer grouping of identical objects. This physical setup remains consistent across all three experimental protocols to ensure uniform evaluation. Further details on the baseline implementation and specific setup configurations will be provided in subsequent sections.
\vspace{-0.3cm}
\subsection{OPO Protocol}
Only-pick-once (OPO) protocol is designed to reflect and standardize the grasping process of picking tasks where picking up an exact number of items is the goal. For example, in warehouses, workers need to fulfill an order of multiple identical items, such as cosmetic jars, USB drives, or similar small components. 

\paragraph{Goal}
The objective of the OPO protocol is to measure the accuracy and efficiency of the evaluated approach while only allowing the robot to perform one round of picking, which includes approaching, forming pre-grasp, grasping/re-grasping, and lifting. It directly measures how well a robotic system can grasp a specific number of objects in a single round of picking. 

\paragraph{Procedure}
\begin{itemize}
\item Provide a target number p to the robotic system, start the timer, and record $t_0$.
\item When the approaching ends, record $t_1$, and report the approaching time $t_a=t_1-t_0$.
\item When the robotic arm starts to rise, record $t_2$, and report the grasping/regrasping time $t_g=t_2-t_1$.
\item When the robotic arm stops rising, record $t_3$, and report the lifting time $t_l = t_3-t_2$.
\item Wait three seconds.
\item Exam how many objects are in the robotic hand, report it as $q$.
\item Repeat 100 times.
\end{itemize}
The total time cost of finishing one OPO procedure is $t_{OPO} = t_a+t_g+t_l$. 

\vspace{-0.3cm}
\subsection{APT Protocol}
Accurate pick-transferring (APT) protocol represents a frequent and practical task in warehouse automation and logistics environments, designed to fulfill requirements where the target quantity of objects exceeds the maximum capacity of a robot's gripper in a single grasp. For instance, imagine a situation where a robot is assigned to move a certain number of apples into a crate for shipping. Since the robotic hand can only grasp several apples at once, the robot must repeat the process a few times to reach the desired amount.

\paragraph{Goal}
The APT protocol measures the efficiency of a robotic system in grasping and transferring objects across multiple attempts, allowing for sequential picks when the object count exceeds the gripper’s capacity. It evaluates the system’s ability to plan and execute actions for each grasp, ensuring efficiency. Essentially, it extends the OPO protocol by incorporating multiple rounds of picking to achieve the goal.

\paragraph{Procedure}
\begin{itemize}
\item Provide a target number $N_{target}$ (usually more than 5) to the robotic system.
\item Select a target number $p$ of objects to grasp for the current round. 
\item Perform an OPO round and record the $t_{OPO}$ and the number of objects grasped $q$.
\item Transfer the objects to the target bin and record the transfer time $t_{transfer}$ (including the return to the start position for the next round).
\item Update the target with $N_{target}$ - $q$.
\item Repeat this process until the target is zero. Record the number of performed OPO procedures $m_{OPO}$
\item Repeat 100 times.
\end{itemize}

In this procedure, if the number of objects in the target bin exceeds the intended amount, the excess objects must be picked up from the target bin and returned to their original bin. The additional time taken for this step should also be included. The total time cost of finishing one APT procedure is $t_{APT}= m_{OPO}(t_{OPO}+t_{transfer})$. If, at a particular round, the $p=1$, an SOP procedure will be performed and the time cost will be assigned to $t_{OPO}$ for that round. 

\vspace{-0.3cm}
\subsection{PTA Protocol}
Pick-transferring-all (PTA) protocol represents a highly challenging task that demands precise planning and adaptability to handle difficult grasping scenarios. This task is particularly complex when the system must clear objects with similar shapes and sizes, which can create additional challenges such as dense packing, limited distinguishability, or unpredictable object interactions. For example, consider a scenario in a warehouse where a robot is tasked with clearing a bin filled with uniformly shaped and sized plastic cups stacked tightly together. The robot must plan its actions to efficiently remove all the cups while addressing potential difficulties such as cups slipping, deforming, or sticking to one another due to friction or static.

\paragraph{Goal}
The primary goal of the PTA protocol is to measure the efficiency of grasping and transferring all objects from the source bin to the target bin through a series of picks and transfers, aiming to optimize operational efficiency and minimize handling time. This protocol evaluates the system’s ability to manage a complete clearing task, requiring effective planning and execution across multiple grasping attempts. By integrating aspects of both OPO (focusing on precise object counts) and APT (handling larger quantities across several attempts), the PTA protocol challenges the robot’s capacity for sequential task planning, adaptability, and consistent performance in dynamic environments.

\paragraph{Procedure}
\begin{itemize}
\item Select a target number $p$ of objects to grasp for the current round. 
\item Perform an OPO round and record the $t_{OPO}$ and the number of objects grasped $q$.
\item Transfer the objects to the target bin and record the transfer time $t_{transfer}$.
\item Repeat this process until the original bin is empty. Record the number of performed OPO procedures $m_{OPO}$
\item Repeat 100 times.
\end{itemize}
The total time cost of finishing one PTA procedure is $t_{PTA}= m_{OPO}(t_{OPO}+t_{transfer})$. If, at a particular round, the $p=1$, an SOP procedure will be performed and the time cost will be assigned to $t_{OPO}$ for that round. 

\vspace{-0.3cm}
\subsection{Evaluation Metrics}
We propose several evaluation metrics using the recorded values in the protocol procedures to comprehensively assess the accuracy, reliability, and efficiency of MOG techniques. They are picking accuracy (PA), overall success rate (OSR), availability rate (AR), and cost of grasping per unit (CGPU). 

\emph{Picking Accuracy (PA)} is the normalized root mean square error (RMSE). It measures the accuracy of a picking outcome compared to the desired, making it particularly suitable for evaluating grasping performance in OPO Protocol. We define PA in Equation \ref{PA}, where $N_{total}$ is the total number of grasp trials performed, $target$ is the desired number of objects to be grasped, and $o_i$ is the number of objects grasped in the $i_{th}$ grasping trial. It serves as a performance indicator, particularly for tasks that can accommodate minor errors—such as grabbing a few ice cubes to toss into a drink. It can also be applied to analyze how performance impacts subsequent tasks, as measured by APT.
\vspace{-0.3cm}
\begin{equation} \label{PA}
PA(object, target) = \sqrt{\frac{\sum_{n=1}^{N_{total}} (o_i - target)^2}{N_{total}}},
\end{equation}


\emph{Overall Success Rate (OSR)} is an intuitive and widely adopted metric in robotic grasping evaluations, representing the proportion of successful grasps relative to the total number of attempts. It is defined in Equation \ref{OSR}, where $N_{success}$ is the number of grasp trials achieving the target. It is also a good measurement for evaluating grasping performance in OPO Protocol. It serves as a performance indicator, particularly for tasks that can high degree of accuracy is critical - such as order fulfillment. As PA, it can also be applied to analyze how performance impacts subsequent tasks, as measured by APT.
\vspace{-0.2cm}
\begin{equation} \label{OSR}
OSR(object, target) = \frac{N_{success}}{N_{total}},
\end{equation}


\emph{Cost of Grasping per Unit (CGPU)} is designed to measure the efficiency of an MOG process. It directly compares the MOG process with the SOG process. In this metric, we assume the SOG process is always perfect, and it takes exactly the number of picking-transferring as the targeted number of objects, $k$. CGPU can be calculated using the average $t_{PTA}$ or $t_{APT}$ normalized by the time of applying single-object picking and transferring ($t_{sopt}$). 
\vspace{-0.3cm}
\begin{equation} 
\begin{split}
CGPU = \frac{t_{PTA/APT}}{k t_{sopt}}
\end{split}
\end{equation}

If we assume each round of OPO procedure uses the same time as a round of SOP, the CGPU can be simplied as 
\begin{equation} \label{CGPU-lift}
CGPU_{s} = \frac{N_{OPO}}{k},
\end{equation}
where $N_{OPO}$ represents the total OPO rounds the system executes for PTA or APT, including both successful and failed grasping trials, to obtain the target number of $k$ objects. For example, for picking and transferring five apples using MOG, the PTA procedure has two rounds of OPOs. $s$ is short for simplified. The $CGPU_{s} = \frac{2}{5}$. 

Since there could be overhead in each OPO process caused by extra steps such as regrasping introduced in \cite{10502193} to improve PA. Therefore, the CGPU contains 
\vspace{-0.3cm}
\begin{equation} \label{CGPU}
\begin{split}
CGPU =  CGPU_{s} + \lambda_{o} CGPU_{s}
\end{split}
\end{equation}

$CGPU_{o}$ represents the overhead in the MOG process that is not in the SOP process. The average extra cost for each picking-transferring measured $t_{PTA}$ or $t_{PTA}$ compared to SOP is normalized by the time SOP and defined as $\lambda_{o}$. 

However, since $t_{PTA}$ and $t_{PTA}$ contain transferring time $t_{transfer}$, it could be affected by the setup. To remove the variances in the setup, we define the adjusted CGPU as 
\vspace{-0.1cm}
{\small
\begin{equation} \label{CGPU}
\begin{split}
CGPU_{a} =  (1+ \lambda_{ao}) CGPU_{s}
\end{split}
\end{equation}}
where $\lambda_{ao} = \frac{t_{OPO}-t_{SOG}}{t_{SOG}}$.

In certain scenarios, it is important to evaluate the system's capability to effectively identify and group objects for MOG. To address this, we introduce the \emph{Availability Rate (AR)} as a performance metric. AR measures the percentage of arbitrary scenes where the system determines it is feasible to grasp $k$ objects simultaneously. This metric provides insight into the system's ability to locate and group objects based on their spatial arrangement and graspability. A higher AR indicates that the system can efficiently identify opportunities for MOG.

Our chosen metrics build upon and extend those commonly found in SoA benchmarks. Metrics like RMSE and OSR are rooted in traditional single-object grasping evaluations, but their adaptation to the multi-object context sets them apart. For example, benchmarks such as the YCB Object\cite{doi:10.1177/0278364917700714} and the ACRV Picking Benchmark\cite{leitner2016acrvpickingbenchmarkapb} prioritize success rates and object-handling efficiency but often lack metrics that address the cumulative challenges of multi-object scenarios. By introducing $CGPU$ and $CGPU_{a}$, we address a critical gap in SoA benchmarks, enabling a comprehensive assessment of both effectiveness and efficiency in MOG tasks.

\vspace{-0.3cm}

\begin{table}[h!]
\centering
\caption{Summary of MOG Benchmarks}
\label{tab-benchmarking-summary}
\renewcommand{\arraystretch}{3} 
\setlength{\tabcolsep}{2.5pt} 
\begin{tabular}{|p{2cm}|p{2.2cm}|p{2.3cm}|}
\hline
\textbf{MOG Protocol} & 
\textbf{Goal} & 
\textbf{Evaluation Metrics} \\
\hline
OPO & 
Grasp a small, specific target number of objects in a single attempt. & 
OSR, RMSE, AR, CGPU \\
\hline
APT & 
Grasp and transfer a large, specific target number of objects in multiple attempts. & 
$CGPU_{adjusted}$ \\
\hline
PTA & 
Clear all objects from the environment in multiple grasps. &
$CGPU_{adjusted}$ \\
\hline
\end{tabular}
\end{table}
\vspace{-0.7cm}

\subsection{Environment}
For benchmarking MOG, we select six fundamental geometries — sphere, cube, cuboid, cylinder, cone, and ellipsoid, shown in Table \ref{tab:objects}. The selected shapes are convex, without functional elements such as handles or cavities. Objects whose outer shape can be described using one of the fundamental geometries are prevalent in common grasp datasets such as YCB \cite{doi:10.1177/0278364917700714}, ensuring that our selection is aligned with widely used evaluation standards. Additionally, these shapes are commonly encountered in industrial settings, where they appear in products, packaging, and components across sectors like manufacturing, logistics, and assembly. By focusing on these widely used shapes, we ensure relevance to both academic benchmarks and practical, real-world robotic applications.

We also carefully considered the size variations for both simulated and real-world systems to ensure comprehensive evaluation and compatibility. For the simulation experiments, we selected three sizes—small, medium, and large—for each of the basic shapes, enabling us to analyze their performance across a wide range of dimensions. In the real-world system, we narrowed the selection to two distinct sizes for each shape, prioritizing practical considerations while maintaining sufficient variability to validate the system's adaptability and effectiveness. We define small, medium, and large object sizes based on both simulated and real-world examples inspired by the YCB dataset \cite{doi:10.1177/0278364917700714}, ensuring consistency and practical relevance. Simulated objects for each basic shape are categorized into three sizes, with dimensions ranging from approximately 30–40 mm for small to large. Real-world objects matched to these categories extend this range slightly (e.g., spheres from D40 to D52 mm) to incorporate practical variability while maintaining alignment with standardized benchmarks. This approach allowed us to bridge the gap between controlled simulations and real-world applications.

\begin{table*}[t!]
\caption{Objects Used for Grasping}
\label{tab:objects}
\begin{center}
\begin{tabular}
{|p{1.5cm}|p{1.5cm}|p{1.5cm}|p{1.5cm}|p{1.5cm}|p{1.5cm}|p{1.5cm}|p{1.5cm}|p{1.5cm}|}
\hline
\textbf{Basic Shape} & \textbf{picture} & \textbf{size 1(mm)} & \textbf{size 2(mm)} & \textbf{size 3(mm)} & \textbf{real object 1} & \textbf{size(mm)} & \textbf{real object 2} & \textbf{size(mm)} \\ 
\hline
Cube & 
\raisebox{-\totalheight}{\includegraphics[width=1.5cm, height=1.5cm, keepaspectratio]{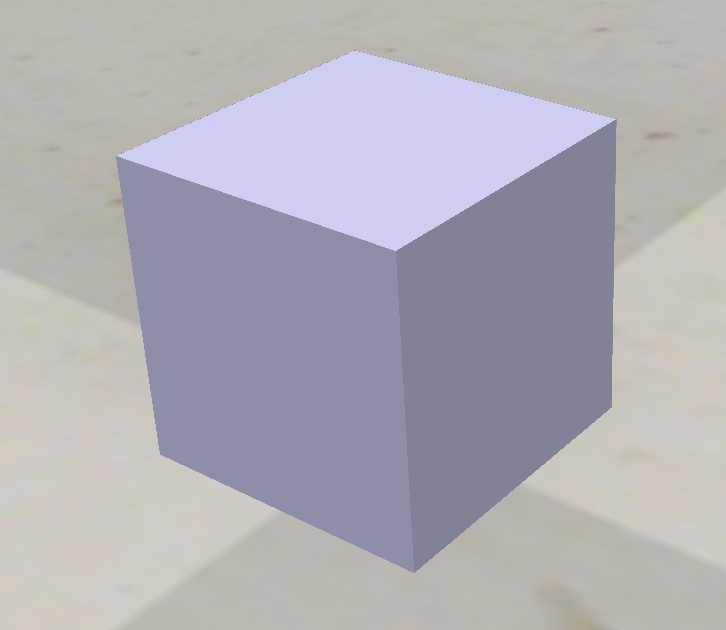}} & 
cube\_s\_s: 30 &
cube\_m\_s: 35 &
cube\_l\_s: 40 &
\raisebox{-\totalheight}{\includegraphics[width=1.5cm, height=1.5cm, keepaspectratio]{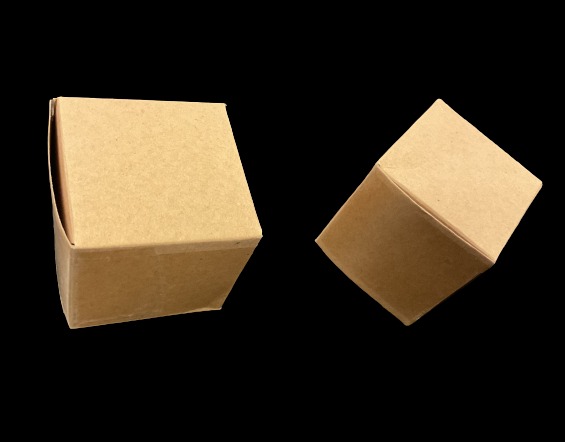}} &
cube\_l\_r: 55 &
\raisebox{-\totalheight}{\includegraphics[width=1.5cm, height=1.5cm, keepaspectratio]{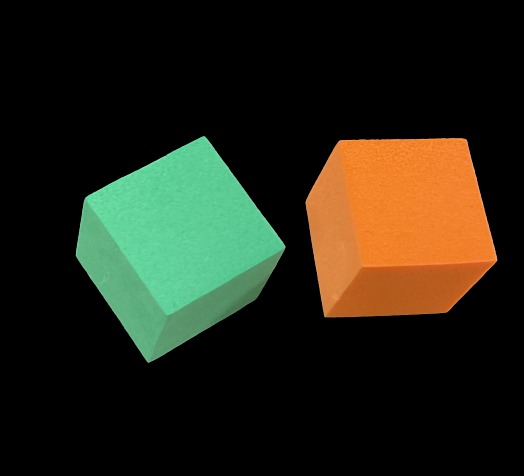}} &
cube\_s\_r: 25\\
\hline

Sphere & 
\raisebox{-\totalheight}{\includegraphics[width=1.5cm, height=1.5cm, keepaspectratio]{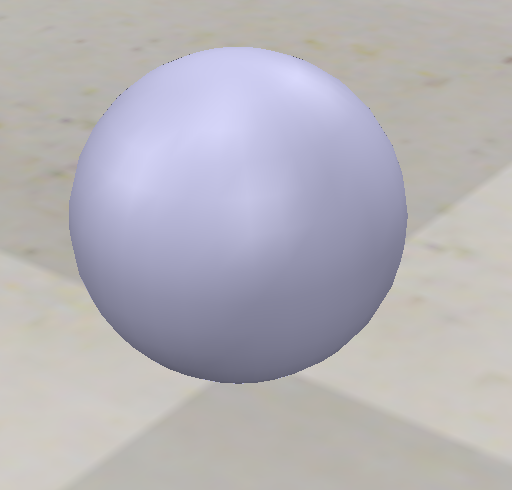}} & 
sphere\_s\_s: D30 &
sphere\_m\_s: D35 &
sphere\_l\_s: D40 &
\raisebox{-\totalheight}{\includegraphics[width=1.5cm, height=1.5cm, keepaspectratio]{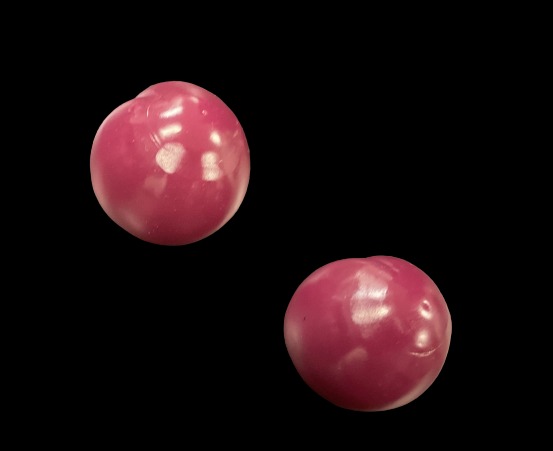}} &
sphere\_l\_r: D52 &
\raisebox{-\totalheight}{\includegraphics[width=1.5cm, height=1.5cm, keepaspectratio]{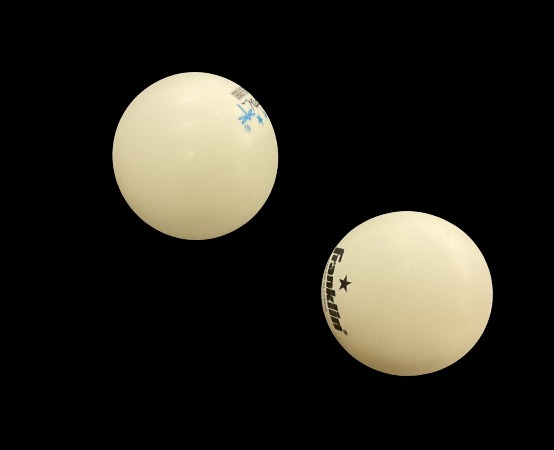}} &
sphere\_s\_r: D40\\
\hline

Cylinder & 
\raisebox{-\totalheight}{\includegraphics[width=1.5cm, height=1.5cm, keepaspectratio]{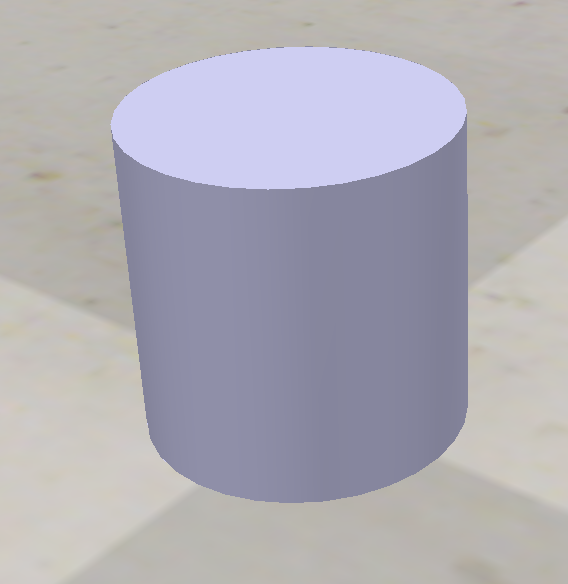}} & 
cylin\_s\_s: D30 H30 &
cylin\_m\_s: D35 H35 &
cylin\_l\_s: D40 H40 &
\raisebox{-\totalheight}{\includegraphics[width=1.5cm, height=1.5cm, keepaspectratio]{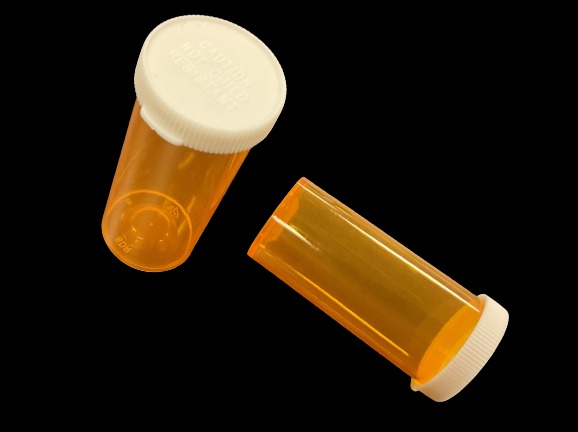}} &
cylin\_l\_r: D26 H63 &
\raisebox{-\totalheight}{\includegraphics[width=1.5cm, height=1.5cm, keepaspectratio]{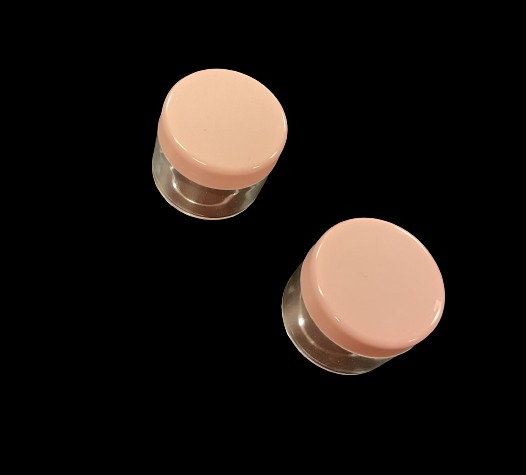}} &
cylin\_s\_r: D36 H30\\
\hline

Cuboid & 
\raisebox{-\totalheight}{\includegraphics[width=1.5cm, height=1.5cm, keepaspectratio]{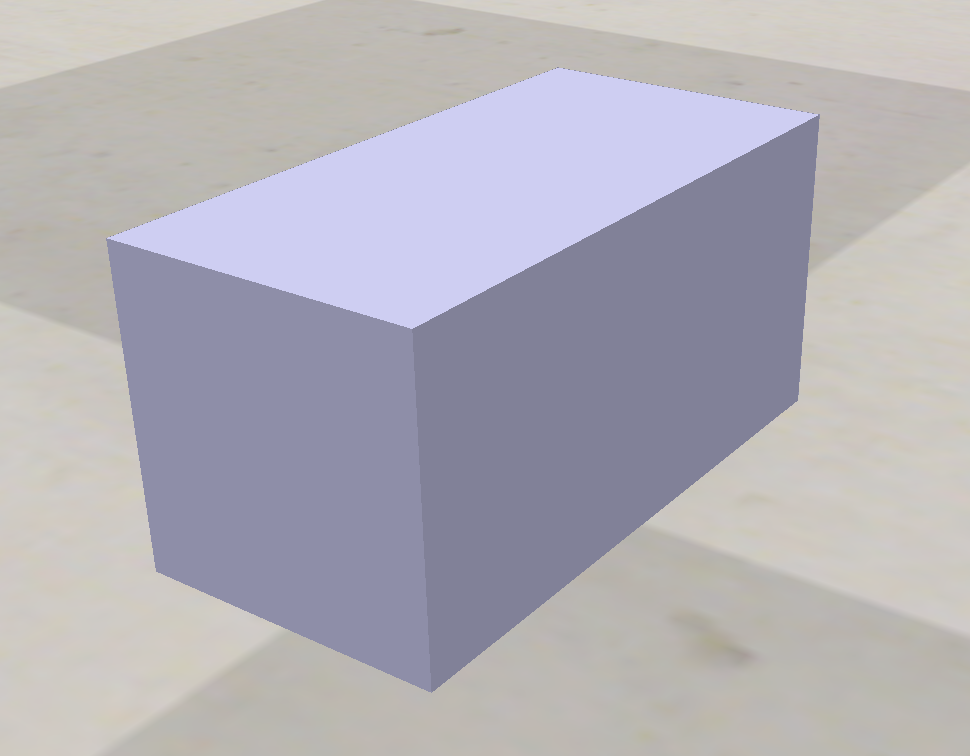}} & 
cuboid\_s\_s: L30 W10 H30&
cuboid\_m\_s: L40 W20 H40 &
cuboid\_l\_s: L50 W30 H50 &
\raisebox{-\totalheight}{\includegraphics[width=1.5cm, height=1.5cm, keepaspectratio]{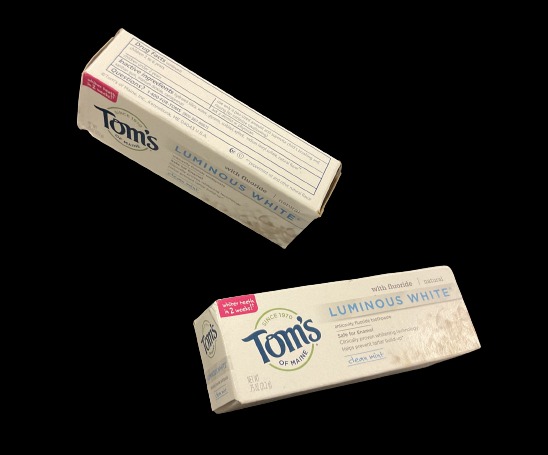}} &
cuboid\_l\_r: L107 W37 H29 &
\raisebox{-\totalheight}{\includegraphics[width=1.5cm, height=1.5cm, keepaspectratio]{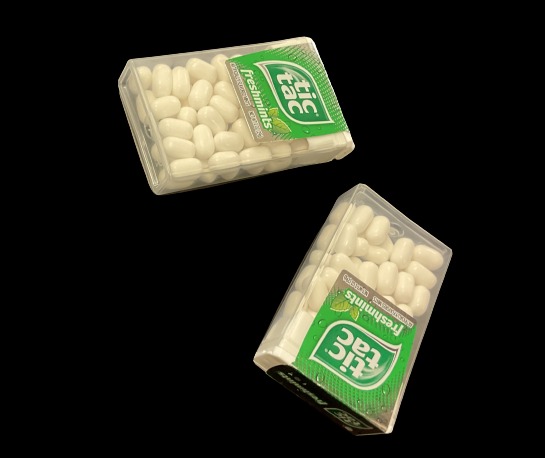}} &
cuboid\_s\_r: L71 W16 H44\\
\hline

Cone & 
\raisebox{-\totalheight}{\includegraphics[width=1.5cm, height=1.5cm, keepaspectratio]{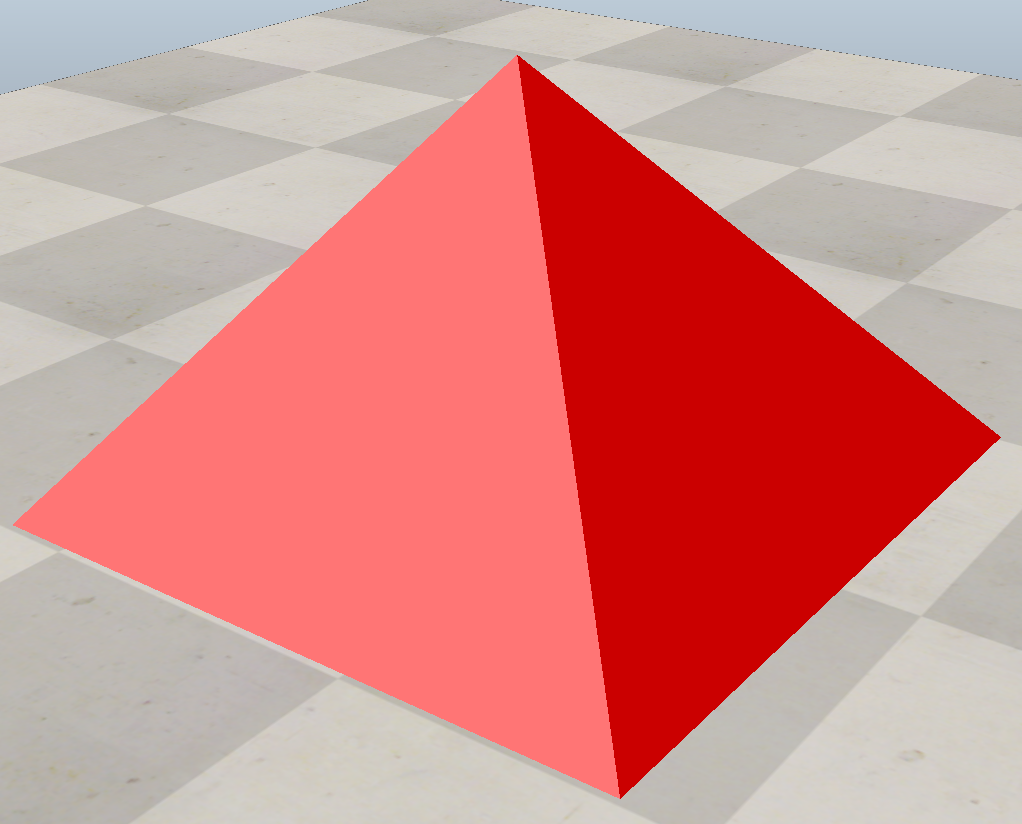}} & 
cone\_s\_s: D10 H30 &
cone\_m\_s: D15 H35 &
cone\_l\_s: D20 H40 &
\raisebox{-\totalheight}{\includegraphics[width=1.5cm, height=1.5cm, keepaspectratio]{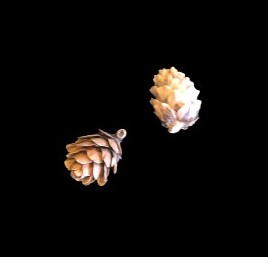}} &
cone\_l\_r: D8 H17 &
\raisebox{-\totalheight}{\includegraphics[width=1.5cm, height=1.5cm, keepaspectratio]{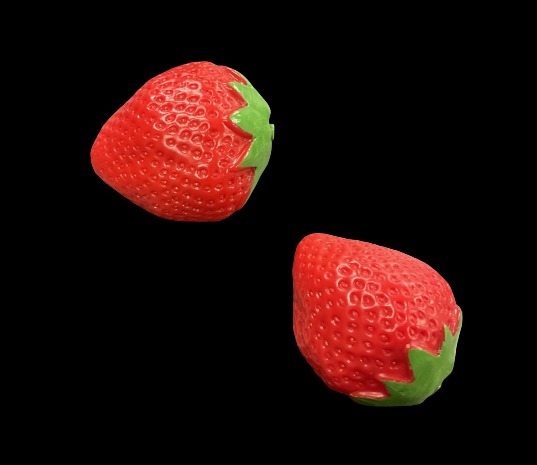}} &
cone\_s\_r: D40 H56\\
\hline

Ellipsoid & 
\raisebox{-\totalheight}{\includegraphics[width=1.5cm, height=1.5cm, keepaspectratio]{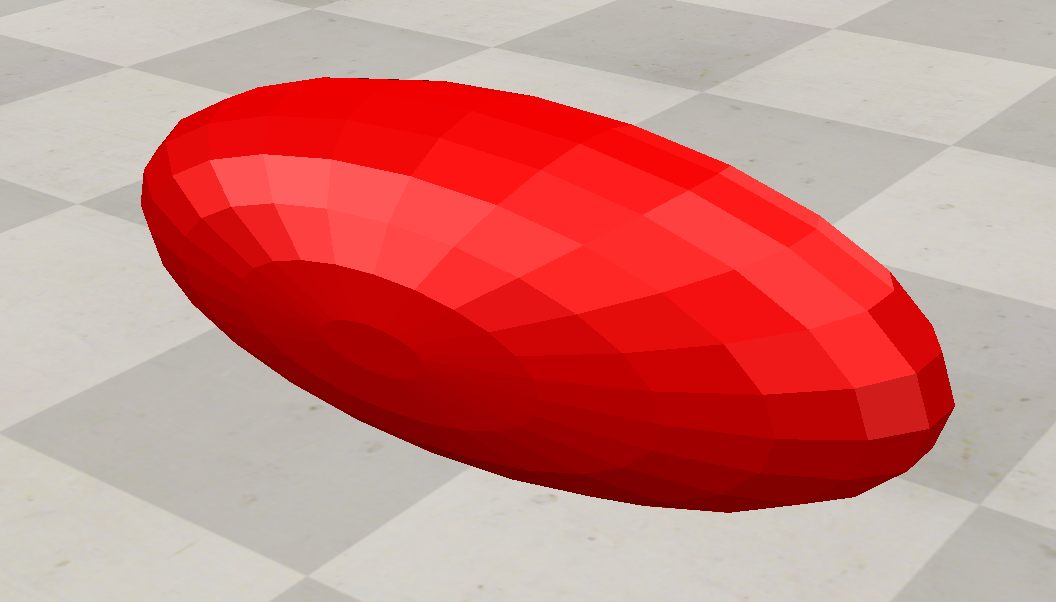}} & 
ellip\_s\_s: L30 S20 &
ellip\_m\_s: L35 S25 &
ellip\_l\_s: L40 S30 &
\raisebox{-\totalheight}{\includegraphics[width=1.5cm, height=1.5cm, keepaspectratio]{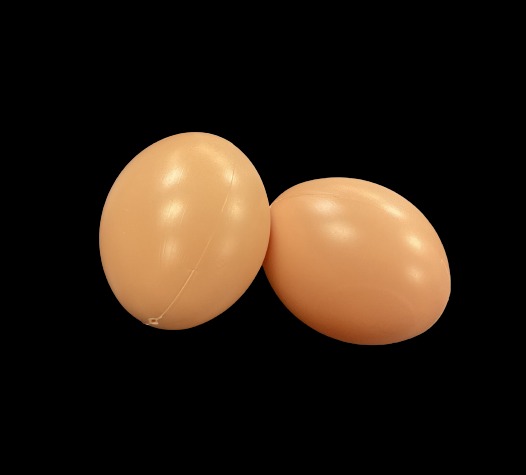}} &
ellip\_l\_r: L60 S45 &
\raisebox{-\totalheight}{\includegraphics[width=1.5cm, height=1.5cm, keepaspectratio]{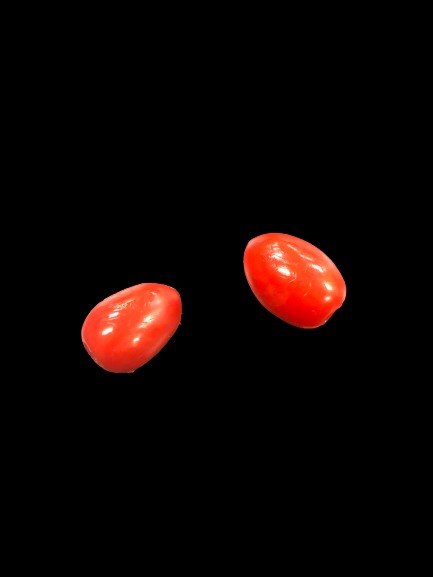}} &
ellip\_s\_r: L33 S21\\
\hline

Irregular & 
\raisebox{-\totalheight}{\includegraphics[width=1.5cm, height=1.5cm, keepaspectratio]{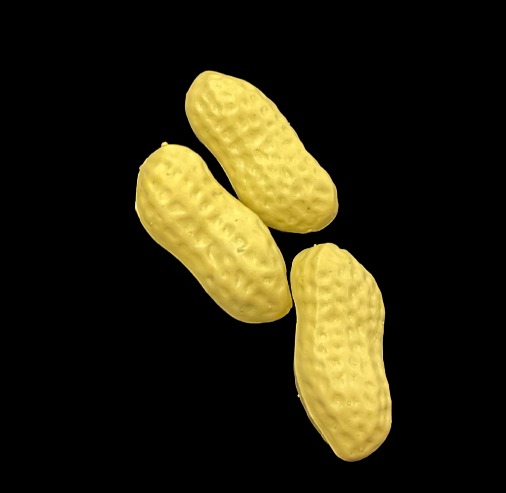}} & 
irreg: L40 D16 & & & & & &\\
\hline

\end{tabular}
\label{table-objects}
\end{center}
\end{table*}

\section{Baselines}
To illustrate how to use the protocols and facilitate comparison, we present the implementation of the protocols on three robotic systems that represent the mainstream pick-and-place setups: parallel gripper, rigid dexterous hand, and soft hand. We carry out the protocols with those robotic systems and our latest algorithms and present the results as baselines. We also carry out the protocols with $5$ human participants and present the human performances and a baseline. 

\begin{figure}[t]
    \centering
    \includegraphics[width=\linewidth]{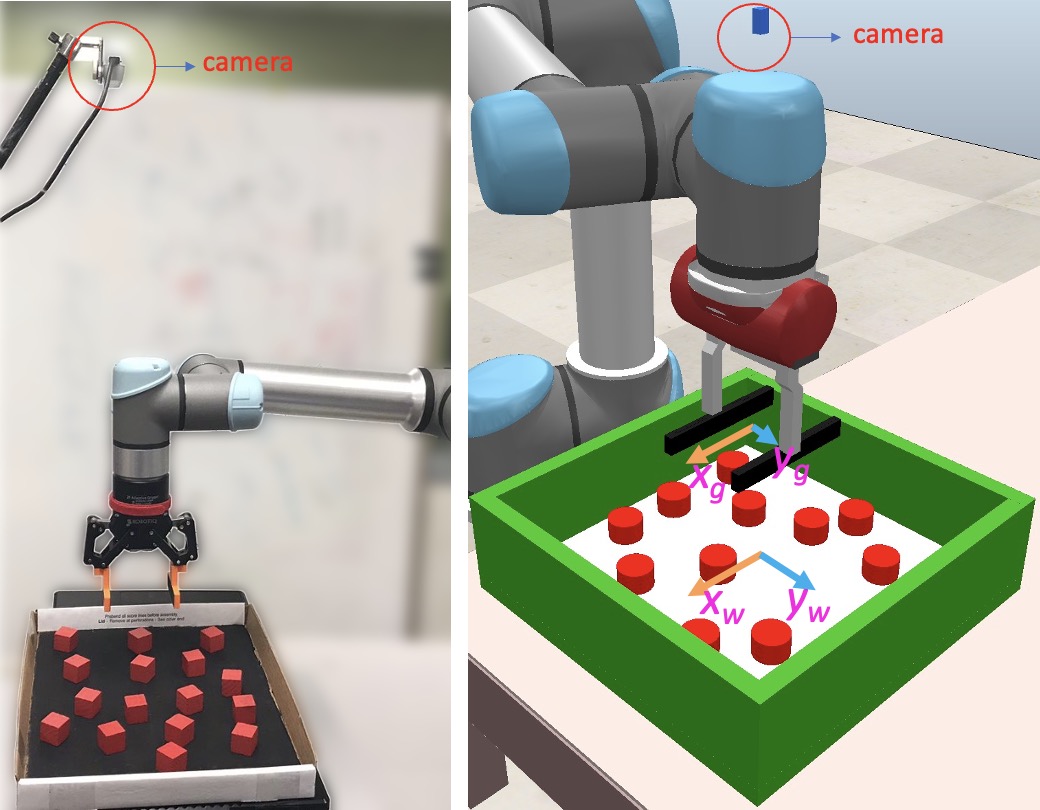}
    \caption{Set up for the Protocol 1 MOG experiments with the Robotiq.}
    \label{robot-setup}
\end{figure}

\paragraph{Setup-Robotiq} 
We set up a typical hand-eye system with a Robotiq 2F-85 gripper, a UR5e arm, and an Intel RealSense Depth Camera D435, widely used in single-object pick and place applications. It is shown in Figure \ref{robot-setup}. We also duplicated the setup in CoppeliaSim for simulation. 

\begin{figure}[t]
    \centering
    \includegraphics[width=\linewidth]{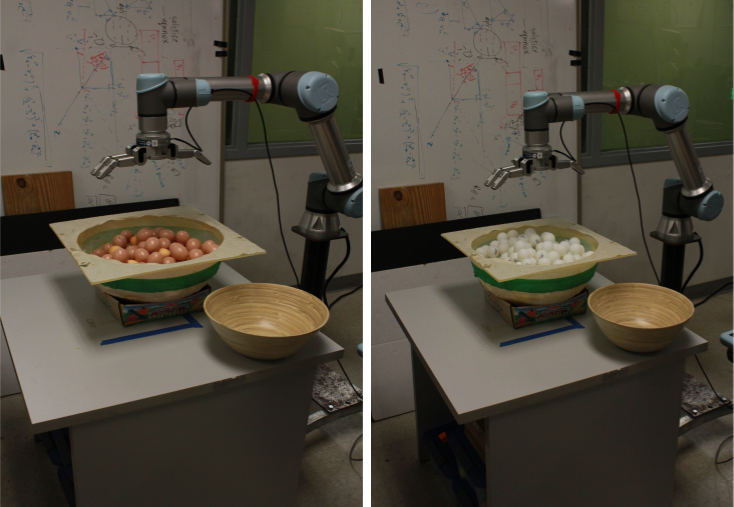}
    \caption{Set up for the Protocols 1 \& 2 MOG experiments with the Barretthand.}
    \label{fig-set_up}
\end{figure}

\paragraph{Setup-Barrett} 
With the same UR5e robotic arm and the camera, we replace the gripper with a Barrett hand as shown in Fig. \ref{fig-set_up}. It represents a traditional grasping and manipulation system with dexterous in-hand manipulation capability. The hand has three fingers and four DoFs. It has tactile sensors on the fingertips and palm to perceive complex grasping patterns in real-time. 

\begin{figure}[t]
    \centering
    \includegraphics[width=\linewidth]{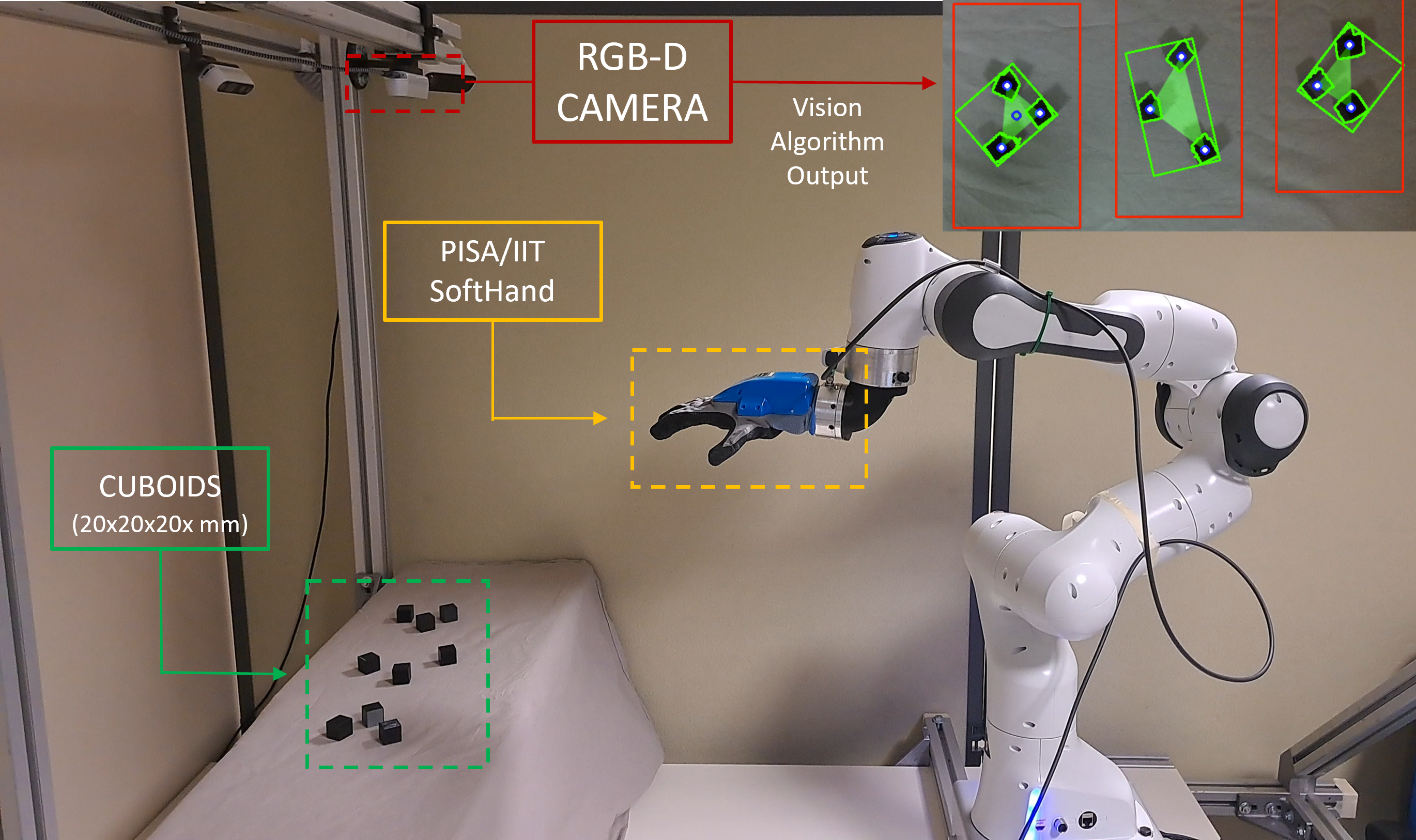}
    \caption{Set up for the Protocol 1 MOG experiments with the PISA/IIT Softhand-2. Objects are placed in clusters on the table in front of the robot, to test the feasibility and repeatability of grasping 3 at a time.}
    \label{fig:setup_SoftHand}
\end{figure}

\paragraph{Setup-SoftHand}
We also implement a robotic system with a PISA/IIT SoftHand-2 mounted on a Franka Emika Robotic, with an Intel RealSense Depth Camera D455, as shown in Figure \ref{fig:setup_SoftHand}. The SoftHand-2 is an anthropomorphic, underactuated soft gripper, featuring 19 DoFs and 2 degrees of actuation, coordinated finger motions called "synergies" enabled by the elastic ligaments between the phalanxes \cite{8373731}. These synergies are designed to imitate the most commonly observed behaviors in human hands: the first synergy realizes the opening and closing of the thumb and fingers, while the second operates a relative rotation between the fingers \cite{Santello10105}.

For the setups employing Robotiq and Barretthand, we directly apply the grasping algorithms proposed in our previous works \cite{10502193, ye2023pick}. For Setup-SoftHand, we developed the following grasping strategy. Since the objects lay on a table surface, the strategy consists of clustering the cubes exploiting the depth information coming from the camera. Constrained K-Means are used to cluster groups of $K$ objects based on the closeness of the objects in the plane. Clusters are passed through a feasibility check to ensure that they can be grasped by the SoftHand-2. Knowing the hand dimensions and the intrinsic parameters of the camera, a rectangle is projected on top of the center of each cluster with an appropriate orientation to check whether the hand is able to cover all the objects within the cluster, without colliding with the other ones. If one of the clusters satisfies this condition, it is selected as the target; otherwise, a cluster of $K$ objects satisfying such conditions must be formed manually by the robotic system. In the latter case, $K$ cubes that are nearest to each other are identified and the system plans three consecutive moves to group them. Once a feasible group of cubes exists, the robot executes a parametrized sequence of Cartesian movements. It first moves on top of the cluster with the hand completely open, parallel to the table, and with proper orientation to maximize the coverage of the cluster, then the hand is pre-configured exploiting its second synergy, to maximize the volume enclosed by the fingers, and moves down to the cluster, stopping a few centimeters from the table. At this point, the robot drags the objects to a predefined position and closes the hand. This strategy is similar to what a person does removing some crumbs from the table.

\begin{figure*}[!htb]
    \centering
    \includegraphics[width=0.85\textwidth]{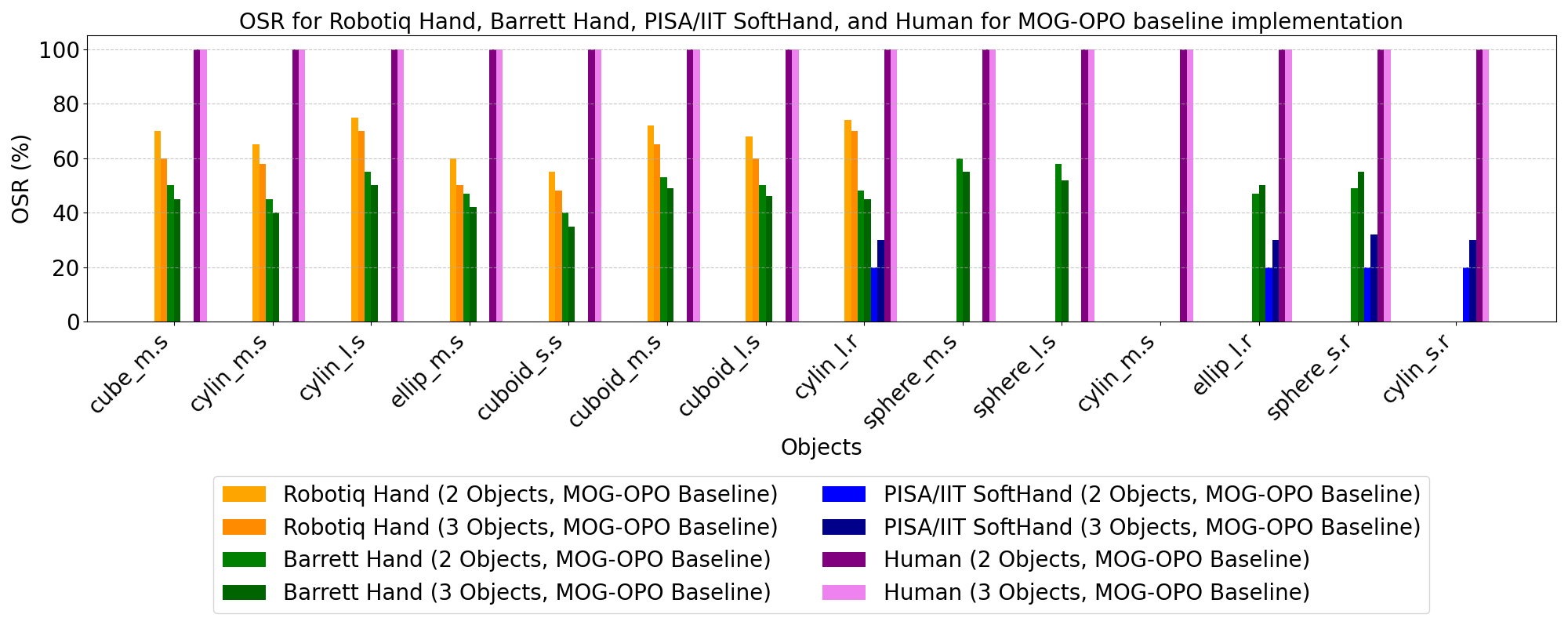}
    \vspace{-0.2cm}
    \caption{OSR for Robotiq Hand, Barrett Hand, PISA/IIT SoftHand-2, and Human Performance when Targeting 2 and 3 Objects.}
    \label{fig: OSR}
\end{figure*}
\vspace{-0.3cm}

\begin{figure}[!htb]
    \centering
    \includegraphics[width=0.48\textwidth]{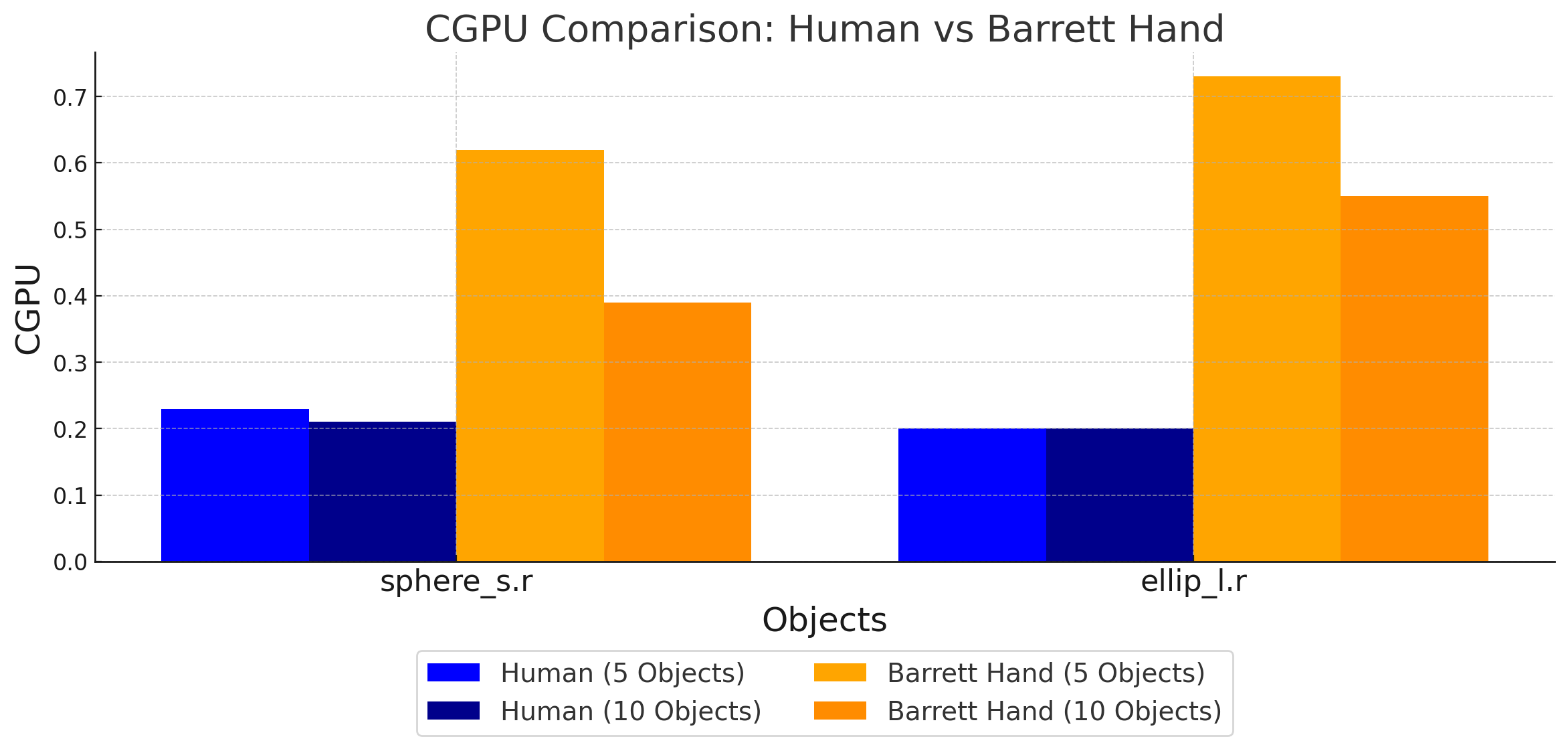}
    \caption{CGPU Comparison Between Human and Barrett Hand for Grasping 5 and 10 Objects.}
    \vspace{-0.2cm}
    \label{fig: CGPU}
\end{figure}
\vspace{-0.3cm}

\vspace{-0.1cm}
\subsection{OPO Baselines}
The results of MOG under Protocol OPO are shown in Fig. \ref{fig: OSR}. In the simulated setup, the Robotiq hand achieved high OSR values when grasping two objects across different object shapes, such as $97.5\%$ for $cube\_m\_s$, $96.5\%$ for $cyli\_m\_s$, and $98.5\%$ for $sphere\_s\_s$. However, OSR values decreased significantly as the target number increased. For example, OSR for $cube\_m\_s$ dropped from $97.5\%$ when grasping 2 objects to $6.5\%$ when grasping 4 objects, reflecting the challenges in maintaining grasp success as the number of objects increases. The results for the Robotiq hand in the real system highlight differences in performance across object numbers. For $cylin\_l\_r$, with OSRs of $81.25\%$ for 2 objects, $40.00\%$ for 3 objects, and only $10.00\%$ for 4 objects. Overall, the Robotiq hand shows strong grasping capabilities for a smaller number of objects but struggles as the task complexity and target numbers increase.

The real setup evaluated OSR and RMSE with the PISA/IIT SoftHand-2 grasping 2 and 3 objects of cube and cylindrical shapes. OSR with 3 objects was greater with $cube\_s\_r$ and equal with $cylin\_s\_r$ with respect to 2 objects. RMSE with 3 $cube\_s\_r$ was higher than using 2 $cube\_s\_r$, while with $cylin\_s\_r$ RMSE achieved a bit less score with 3 objects than considering 2 objects. This can be explained by accounting for the higher intra-forces between the objects in the case of the cubes with respect to the cyilinders when grasped by the compliant hand, which favors successful grasping of a higher object count in the first case with respect to the second. Although its compliant
behavior can prevent the execution of precisely repeatable tasks, SoftHand intrinsic compliance allows
it to adapt to various items and leverage environmental
constraints, fostering MOG applications in heavily
unstructured environments. 

In the simulated setup, the Barrett Hand demonstrated varying levels of OSR and RMSE, with OSR decreasing as the target number increased. For example, the $sphere\_m\_s$ object type achieved an OSR of $63\%$ when grasping $2$ objects, which dropped to $55\%$ for $3$ objects, with corresponding RMSE values of 0.7746 and 0.9798, respectively. Similar trends were observed with other shapes, such as $cuboid\_m\_s$.

In the real setup, OSR values were generally higher for tasks involving $2$ objects. For instance, the $sphere\_m\_r$ object achieved an OSR of $71\%$ with an RMSE of 0.6232 for $2$ objects, compared to $60\%$ with an RMSE of 0.8521 for $3$ objects. These results highlight the system's effectiveness at handling smaller target numbers, with relatively consistent accuracy and precision.


\vspace{-0.3cm}
\subsection{APT Baselines}
The results for Protocol APT, targeting 5 and 10 objects, are shown in Fig. \ref{fig: CGPU}. In the simulated setup, the Barrett Hand's performance declined as the target number of objects increased. For instance, the $sphere\_m\_s$ object type achieved $CGPU_{adjusted}$ values of 0.70 for $5$ objects and 0.52 for $10$ objects. Similar trends were observed with the $cuboid\_m\_s$ object type, which recorded CGPU values of 0.79 for 5 objects and 0.71 for $10$ objects.

In the real setup, $CGPU_{adjusted}$ values also decreased with higher target numbers. For example, $sphere\_m\_r$ achieved a $CGPU_{adjusted}$ of 0.62 for $5$ objects, which dropped to 0.39 for $10$ objects. Similar patterns were observed with the $ellip\_l\_r$ object type, where $CGPU_{adjusted}$ values ranged from 0.61 for $5$ objects to 0.51 for $10$ objects. These results emphasize the challenges of scaling up grasping operations, with performance becoming less efficient as the number of objects increases.

Overall, these findings highlight the limitations of the Barrett Hand in handling higher target numbers under Protocol 2. While the system demonstrates reasonable grasping efficiency for $5$ objects, its performance significantly declines as the number of objects increases to $10$. This trend underscores the growing difficulty of managing larger object groups, particularly in unstructured or cluttered environments, where precise and reliable multi-object grasping becomes increasingly challenging.

\vspace{-0.3cm}
\subsection{PTA Baselines}
For this protocol, we only have human experiments, since it is challenging for the current robotic system to achieve. The setup designed to establish a human baseline consists of two bowls for placing objects and executing the desired tasks. All 13 objects were included in this baseline, and five human subjects performed several trials for each protocol proposed in this paper: OPO with a target of 2, OPO with a target of 3, transferring 5 objects, transferring 10 objects, and clearing the scene. The average hand length of the subjects is 18 cm and the average hand width is 9 cm. In some cases, the setup includes only 10 objects, so the tasks of clearing the scene and transferring 10 objects overlap. The results are presented in Fig. \ref{fig: OSR} and Fig. \ref{fig: CGPU}.
The results of the human experiments serve as a good comparison with robotic performance and can be used as an ideal upper limit to the robot that will be reached or surpassed. For $CGPU$, the better performance results in a score closer to $0$; indeed, the average for the human subjects is much closer than the robot. For example, when transferring $10$ cylinders the simulation gets $0.51$ and $0.69$, while the human results are $0.23$ for the pill container. Even in the worst $CGPU$ for the human experiments ($0.38$ for transferring 10 small cubes), the robotic system is outperformed by more than twice ($0.71$ for transferring medium cubes).

\section{Discussion and Conclusion}
Standardized benchmarking is crucial for evaluating and comparing robotic systems in MOG. Consistent metrics and protocols enhance reproducibility and facilitate meaningful comparisons, driving the development of more robust and versatile robotic grasping systems. In this work, we present three standardized protocols for benchmarking robotic MOG: the OPO Protocol, the APT Protocol, and the PTA Protocol. Each protocol targets a distinct aspect of MOG, evaluating fundamental grasping accuracy, sequential picking and transferring, and large-scale object clearance, respectively. Together, they provide a structured framework for assessing robotic grasping performance across diverse real-world scenarios.

Our protocols address critical shortcomings in prior benchmarks by introducing evaluation metrics tailored to multi-object grasping. The evaluation relies on four key metrics: PA, OSR, CGPU, and AR. The PA metric ensures a fundamental measure of accuracy, while OSR and AR provide reliability assessments across different grasp attempts. CGPU offers a more comprehensive measure of efficiency, capturing the scalability of grasping strategies in cluttered or high-object-count environments. By following these standardized protocols, researchers can objectively compare robotic grasping strategies across different systems, enhancing reproducibility and driving advancements in MOG techniques.

The MOG baseline experiments highlight robotic hands' strengths and limitations across benchmarking protocols. While effective in structured tasks with fewer objects, performance declines with increased clutter and irregularity. Humans, however, maintain consistent efficiency in such scenarios. These findings emphasize the need for advancements beyond hardware, particularly in perception, control, and learning. Vision and depth-sensing struggle with occlusion and segmentation, making sensor fusion with tactile and force feedback crucial. Control strategies require adaptive models for dynamic environments, where compliant hands like SoftHand-2 could be key. Learning approaches must evolve beyond single-object tasks, integrating hierarchical reinforcement learning and domain adaptation. Addressing these challenges will enhance robotic grasping, with our benchmarks guiding future research.

\vspace{-0.3cm}
\bibliographystyle{IEEEtran}
\bibliography{references}

\end{document}